\documentclass{article}
\usepackage[utf8]{inputenc}
\usepackage[T1]{fontenc}
\usepackage[hyphens]{url}
\usepackage{amsmath,amssymb}
\usepackage{graphicx}
\usepackage{subcaption}
\title{Learning gradient-based ICA \\by neurally estimating mutual information}
\author{Hlynur Davíð Hlynsson, Laurenz Wiskott}
\date{February 2019}

\begin{document}

\maketitle

\abstract{ Several methods of estimating the mutual information of random variables have been developed in recent years. They can prove valuable for novel approaches to learning statistically independent features. In this paper, we use one of these methods, a mutual information neural estimation (MINE) network, to present a proof-of-concept of how a neural network can perform linear ICA. We minimize the mutual information, as estimated by a MINE network, between the output units of a differentiable encoder network. This is done by simple alternate optimization of the two networks. The method is shown to get a qualitatively equal solution to FastICA on blind-source-separation of noisy sources.}

\section{Introduction}

Independent component analysis (ICA) aims at estimating unknown \textit{sources} that have been mixed together into an \textit{observation}. The usual assumptions are that the sources are statistically independent and no more than one is Gaussian \cite{jutten2003advances}. The now-cemented metaphor is one of a cocktail party problem: several people (sources) are speaking simultaneously and their speech has been mixed together in a recording (observation). The task is to unmix the recording such that all dialogues can be listened to clearly. 

In linear ICA, we have a data matrix $S$ whose rows are drawn from statistically independent distributions, a mixing matrix A, and an observation matrix $X$:

$$X = AS$$

\noindent and we want to find an unmixing matrix $U$ of $A$ that recovers the sources up to a permutation and scaling:

$$ Y = UX $$

The general non-linear ICA problem is ill-posed \cite{hyvarinen1999nonlinear, darmois1953analyse} as there is an infinite number of solutions if the space of mixing functions is unconstrained. However, post-linear \cite{taleb1999source} (PNL) ICA is solvable. This is a particular case of non-linear ICA where the observations take the form

$$X = f(AS)$$

\noindent where $f$ operates componentwise, i.e. $X_{i, t} = f_i \left( \sum_n^N A_{i, n}S_{n, t}\right) $. The problem is solved efficiently if $f$ is at least approximately invertible \cite{ziehe2003blind} and there are approaches to optimize the problem for non-invertible $f$ as well \cite{ilin2004post}.
For signals with time-structure, however, the problem is not ill-posed even though it is for i.i.d. samples \cite{blaschke2007independent, sprekeler2014extension}. 

%  TODO: Cite myself, somehow (?)        

To frame ICA as an optimization problem, we must find a way to measure the statistical independence of the output components and minimize this quantity. There are two main ways to approach this: either minimize the mutual information between the sources \cite{amari1996new, bell1995non, cardoso1997infomax}, or maximize the sources' non-Gaussianity \cite{hyvarinen2000independent, blaschke2004cubica}. 

There has been a recent interest in combining deep learning with the principles of ICA, usually in an adversarial framework, for example Deep InfoMax (DIM) \cite{hjelm2018learning}, Graph Deep InfoMax \cite{velivckovic2018deep} and Generative adversarial networks \cite{goodfellow2014generative}, which utilize the work of Brakel et al. \cite{brakel2017learning}. Our work is distinct from theirs as we do not rely on adversarial training.

\section{Method}

We train an encoder $E$ to generate an output  $\left(z_1, z_2, \dots, z_M\right)$ such that any one of the output components is statistically independent of the union of the others, i.e. $P(z_i, \boldsymbol{z_{-i}}) = P(z_i)P(\boldsymbol{z_{-i}})$, where $$\boldsymbol{z_{-i}} := \left(z_1, \dots, z_{i-1}, z_{i+1}, \dots, z_M\right)$$ The statistical independence of $z_i$ and  $\boldsymbol{z_{-i}}$ can be maximized by minimizing their mutual information 
\begin{equation} \label{mi_definition}
I\left(Z_i; \boldsymbol{Z_{-i}} \right) =
\int_{z} \int_{\boldsymbol{z_{-i}}} 
p(z_i, \boldsymbol{z_{-i}}) \log \left( 
\frac{p(z_i, \boldsymbol{z_{-i}})}{p(z_i)p(\boldsymbol{z_{-i}})} \right) dz_i d\boldsymbol{z_{-i}} 
\end{equation}

 This quantity is hard  to estimate, particularly for high-dimensional data. We therefore estimate the lower bound of Eq.\ (\ref{mi_definition}) using a mutual information neural estimation (MINE) network $M$   \cite{belghazi2018mine}: 
\begin{equation} \label{mine_objective}
I\left(Z_i; \boldsymbol{Z_{-i}} \right) \geq L_i = \mathbb{E}_{\mathbb{J}} \left[M \left( z_i, \boldsymbol{z_{-i}} \right) \right] - \log \left( \mathbb{E}_{\mathbb{M}} \left[ e^{M \left( z_i, \boldsymbol{z_{-i}} \right) } \right]  \right)
\end{equation}

\noindent where $\mathbb{J}$ indicates that the expected value is taken over the   joint and similarly $\mathbb{M}$ for the product of marginals. The networks $E$ and $M$ are parameterized by $\theta_E$ and $\theta_M$. The encoder takes the observations as input and the MINE network takes the output of the encoder as an input.

The E network minimizes $L := \sum_i L_i$ in order for the outputs to have low mutual information and therefore be statistically independent. In order to get a faithful estimation of the lower bound of the mutual information, the  M network maximizes $L$. Thus, in a push-pull fashion the system as a whole converges to independent output components of the encoder network E. 
In practice, rather than training the E and M networks simultaneously it proved useful to train M from scratch for a few iterations after each iteration of training E, since the loss functions of E and M are at odds with each other. When the encoder is trained, the MINE network's parameters are frozen and \textit{vice versa.}

\begin{figure}[h!] 
\centering
 \captionsetup{width=.8\linewidth}
  \resizebox*{0.99   \textwidth}{!}{\includegraphics
{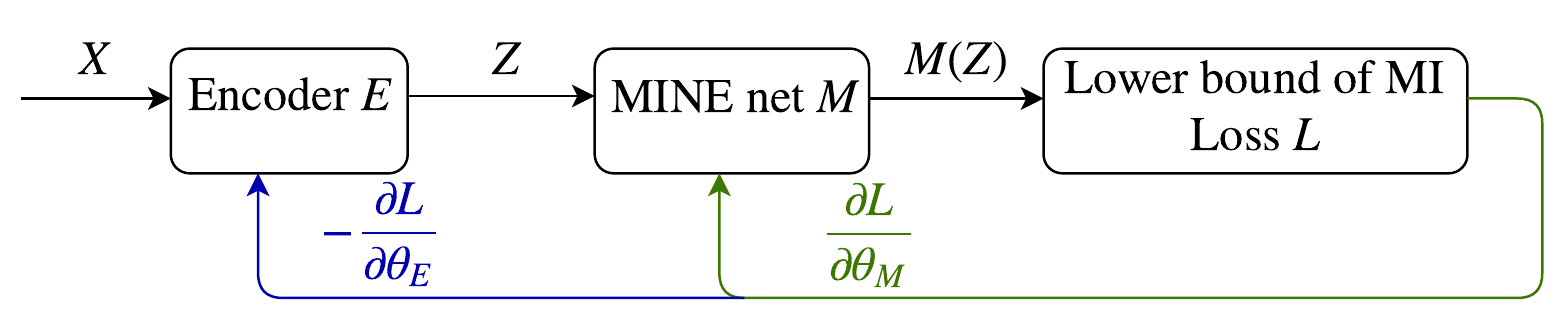}}
\caption{The system learns statistically independent outputs by alternate optimization of an encoder E and a MINE network M parameterized by $\theta_E$ and $\theta_M$. The MINE objective (Eq.\ \ref{mine_objective}) is minimized with respect to $\theta_E$ for weight updates of the encoder but it is \textit{maximized} with respect to $\theta_M$ for weight updates of the MINE network.}
\label{our_method}
\end{figure}

\section{Results}

\begin{figure}
\centering
 \captionsetup{width=.88\linewidth}
\begin{subfigure}[b]{.45\linewidth}
\includegraphics[width=\linewidth]{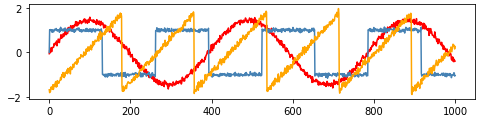}
\caption{The original sources.}\label{sources}
\end{subfigure}
\begin{subfigure}[b]{.45\linewidth}
\includegraphics[width=\linewidth]{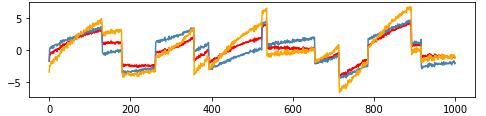}
\caption{Linear mixture of sources.}\label{mixed}
\end{subfigure}

\begin{subfigure}[b]{.45\linewidth}
\includegraphics[width=\linewidth]{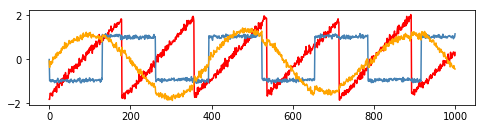}
\caption{Sources recovered by our method.}\label{ours_bss}
\end{subfigure}
\begin{subfigure}[b]{.45\linewidth}
\includegraphics[width=\linewidth]{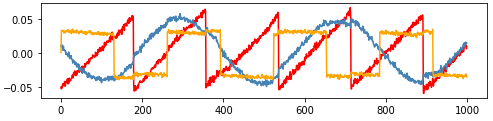}
\caption{Sources recovered by FastICA.}\label{fastica_bss}
\end{subfigure}
\caption{Three independent, noisy sources (a) are mixed linearly (b). Our method recovers them (c) to the same extent as FastICA (d).}
\label{fig:animals}
\end{figure}

We validate the method\footnote{Full code for the results is available at \url{github.com/wiskott-lab/gradient-based-ica/blob/master/bss3.ipynb}} for linear noisy ICA example \cite{sklearn}.
Three independent, noisy sources --- sine wave, square wave and sawtooth signal (Fig.\ \ref{sources}) --- are mixed linearly (Fig.\ \ref{mixed}): 

$$ Y =   \begin{bmatrix}
    1 & 1 & 1 \\
    0.5 & 2 & 1\\
    1.5 & 1 & 2 
  \end{bmatrix} S$$
  \\
The encoder is a single layer neural network with linear activation with a differentiable whitening layer \cite{schuler2018gradient} before the output. The whitening layer is a key component for performing successful blind source separation for our method. Statistically independent random variables are necessarily uncorrelated, so whitening the output by construction beforehand simplifies the optimization problem significantly. 

The MINE network M is a seven-layer neural network. Each layer but the last one has 64 units with a rectified linear activation function. Each training epoch of the encoder is followed by seven training epochs of M. Estimating the exact mutual information is not essential, so few  iterations suffice for a good gradient direction.

Since the MINE network is applied to each component individually, to estimate mutual information (Eq.\ \ref{mine_objective}), we need to pass each sample through the MINE network $N$ times --- once for each component. Equivalently, one could conceptualize this as having $N$ copies of the MINE network and feeding the samples to it in parallel, with different components singled out. Thus, for sample $(z_1, z_2, \dots, z_N)$ we feed in $(z_i ; z_{-i})$, for each $i$. Both networks are optimized using Nesterov momentum ADAM \cite{dozat2016incorporating} with a learning rate of $0.005$. 

For this simple example, our method (Fig.\ \ref{ours_bss}) is equivalently good at unmixing the signals as FastICA (Fig.\ \ref{fastica_bss}), albeit slower. Note that, in general, the sources can only be recovered up to permutation and scaling.

\section{Summary} 

We've introduced a proof-of-concept implementation for training a differentiable function for performing ICA. The method consists of alternating the optimization of an encoder and a neural mutual-information neural estimation (MINE) network. The mutual information estimate between each encoder output and the union of the others is minimized with respect to the encoder's parameters. 
Although this work is in a very preliminary stage, further investigation into the method is warranted. The general nonlinear ICA problem is ill-posed, but it is an interesting question whether this method can work for non-linear problems with low complexity. We can constrain the expresiveness of our encoder by limiting for example the number of layers or number of hidden units in the neural network, thus constraining the solution space of the method. The method is also trivially extended for over- or undercomplete ICA by changing the number of output units. Higher dimensional and real-world data can also be tested. 

As this method can be used for general neural network training, it should be investigated whether useful representations can be learned while solving the ICA task. This method blends nicely into deep learning architectures and the  MINE loss term can be added as a regularizer to other loss functions.   We imagine that this can be helpful for methods such as deep sparse coding to enforce independence between features and disentangle factors of variation.

\bibliography{main}
\bibliographystyle{ieeetr}

\end{document}